\documentclass{article}

\usepackage{arxiv}

\usepackage[utf8]{inputenc} 
\usepackage[T1]{fontenc}    
\usepackage{hyperref}       
\usepackage{url}            
\usepackage{booktabs}       
\usepackage{amsfonts}       
\usepackage{nicefrac}       
\usepackage{microtype}      
\usepackage{lipsum}
\usepackage{graphicx}
\usepackage{multirow}
\usepackage{graphicx}
\usepackage{booktabs}
\usepackage{subcaption}
\usepackage{enumitem}
\graphicspath{ {./images/} }

\title{Identifying and Mitigating Gender Cues in Academic Recommendation Letters: An Interpretability Case Study}

\author{
 Charlotte S. Alexander \\
  Scheller College of Business\\
  Georgia Institute of Technology\\
  Atlanta, GA 30308, USA \\
  \texttt{charlotte.alexander@scheller.gatech.edu} \\
   \And
 Shane Storks \\
  Weinberg Institute for Cognitive Science\\
  University of Michigan - Ann Arbor\\
  Ann Arbor, MI 48109, USA \\
  \texttt{sstorks@umich.edu} \\
  \And
 Souradip Pal \\
  School of Electrical and Computer Engineering\\
  Purdue University\\
  West Lafayette, IN 47906, USA \\
  \texttt{pal43@purdue.edu} \\
  \And
 Sayak Chakrabarty \\
  Department of Computer Science\\
  Northwestern University\\
  Evanston, IL 60208, USA \\
  \texttt{sayakchakrabarty2025@u.northwestern.edu} \\
  \And
 Arushi Sharma \\
  Department of Computer Science\\
  Iowa State University\\
  Ames, IA 50011, USA \\
  \texttt{arushi17@iastate.edu} \\
  \And
 Mlen-Too Wesley \\
  College of Computing\\
  Georgia Institute of Technology\\
  Atlanta, GA 30332, USA \\
  \texttt{mwesley32@gatech.edu} \\
  \And
 Bailey Russo \\
  College of Computing\\
  Georgia Institute of Technology\\
  Atlanta, GA 30332, USA \\
  \texttt{brusso6@gatech.edu} \\
}
\date{}
\begin{document}
\maketitle
\begin{abstract}
Letters of recommendation (LoRs) can carry patterns of implicitly gendered language that can inadvertently influence downstream decisions, e.g. in hiring and admissions. In this work, we investigate the extent to which Transformer-based encoder models as well as Large Language Models (LLMs) can infer the gender of applicants in academic LoRs submitted to an U.S. medical-residency program after explicit identifiers like names and pronouns are de-gendered. While using three models (DistilBERT, RoBERTa, and Llama 2) to classify the gender of anonymized and de-gendered LoRs, significant gender leakage was observed as evident from up to 68\% classification accuracy. Text interpretation methods, like TF-IDF and SHAP, demonstrate that certain linguistic patterns are strong proxies for gender, e.g. ``\textit{emotional}'' and ``\textit{humanitarian}'' are commonly associated with LoRs from female applicants. As an experiment in creating truly gender-neutral LoRs, these implicit gender cues were remove resulting in a drop of up to 5.5\% accuracy and 2.7\% macro $F_1$ score on re-training the classifiers. However, applicant gender prediction still remains better than chance. In this case study, our findings highlight that 1) LoRs contain gender-identifying cues that are hard to remove and may activate bias in decision-making and 2) while our technical framework may be a concrete step toward fairer academic and professional evaluations, future work is needed to interrogate the role that gender plays in LoR review. Taken together, our findings motivate upstream auditing of evaluative text in real-world academic letters of recommendation as a necessary complement to model-level fairness interventions.
\end{abstract}


\section{Introduction}

Natural language processing (NLP) techniques have become central to uncovering subtle patterns of bias in institutional texts. In the realm of academic selection, letters of recommendation (LoRs) are both ubiquitous and influential, yet their narrative nature leaves ample room for implicit stereotyping. Prior work has demonstrated the limits of removing \textit{explicit} cues of gender (e.g., applicant name and pronouns) from hiring materials as a blinding strategy, due to the existence of pervasive \textit{implicit} gendered language \cite{dastinInsightAmazonScraps2018,alexander2022text}. As implicit gender cues can systematically impact hiring decisions \cite{Rice2016HiringDT}, we present a focused \textit{case study} on detecting gendered language in LoRs submitted to a large U.S.\ medical residency program.

Digital submission platforms now archive tens of thousands of LoRs annually, providing an unprecedented opportunity to examine gendered language at scale.  Whereas prior bias audits largely relied on hand-coding a few hundred documents, modern transformer-based language models make it feasible to process entire institutional corpora.  Harnessing this development, we cast gender inference as a text-classification problem and ask: \emph{How much gender signal persists once explicit gender identifiers are removed, and is it possible to remove this implicit gender signal at the dataset-level?}

\paragraph{Case-study framework.}
Our experimental pipeline follows three stages: data curation, model training, and interpretability-driven audit.

\begin{enumerate}
    \item \textbf{Data construction.}
          A corpus of 8,992 LoRs was compiled by annotating applicant gender from official LoR records, and extracting rich metadata and textual features.
    \item \textbf{Attention-based classification.}  
          DistilBERT, RoBERTa, and Llama-2 classifiers are fine-tuned on (i) raw LoRs (non-EDG) and (ii) \emph{explicitly de-gendered} LoRs (EDG) in which pronouns, titles, and kinship terms are neutralized.
    \item \textbf{Interpretability and re-evaluation.}  
          Token-level importance scores (TF-IDF, SHAP) highlight residual implicit gender cues; removing the most predictive cues yields a third training pass of \emph{implicitly de-gendered} LoRs (EDG w/o SHAP Tokens and EDG w/o TF-IDF Tokens), allowing us to quantify \emph{implicit} de-gendering efficacy.
\end{enumerate}

In summary, this case study offers three main contributions: (i) a reproducible framework for detecting gendered language in LoRs, complete with data-processing scripts and an attention-based classification baseline. (ii) documentation and collection/annotation protocols to derive explicitly and implicitly de-gendered texts conducive to further socio-linguistic inquiry and (iii) a critical discussion of technical, ethical, and deployment considerations for AI-assisted admissions workflows, emphasizing transparency, interpretability, and iterative bias mitigation.

\section{Related Works}
Prior work has long gathered empirical evidence of significant gender bias in professional contexts, such as hiring decisions \cite{Isaac2009InterventionsTA,koch2015meta,Rice2016HiringDT,hoover2019powerless,keck2020decoy}.
In the context of open-source software development, \cite{imtiaz2019investigating} found that women's GitHub pull requests were, on average, accepted more frequently than men's, unless the contributor's gender was publicly identifiable, at which point acceptance rates fell significantly. The authors inferred that lower acceptance rates stem not from inferior code quality, but from bias activated by visible gender markers. Similarly, \cite{SIMON2023103423} found systematic differences in patterns of language used in LinkedIn profiles by gender.
AI language model-based text classifiers are effective tools to expose correlations between text data and various class labels and categories \cite{schwartz-etal-2017-effect,gururangan-etal-2018-annotation,poliak-etal-2018-hypothesis,niven-kao-2019-probing}. 
As such, they can identify implicit gender cues in application materials, and when used as hiring tools, they can leverage such cues to inadvertently perpetuate historical hiring bias reflected in training data 
\cite{dastinInsightAmazonScraps2018}.
Highly relevant to our work, \cite{liu-etal-2022-assessing} used a large language model (LLM) to assess gender bias in human-written feedback for medical students, finding that terms related to family and children were more likely to be used in evaluating female students.
These findings support our concern that human-written text can encode latent gender information undetected by na{\"i}ve anonymization.

Meanwhile, NLP technologies like LLMs themselves often learn and reproduce societal stereotypes present in text corpora used to pre-train them, exacerbating concerns of bias.
Prior work has found evidence of gender and other social bias across broad domains and tasks, including in distributional semantic representations \cite{10.5555/3157382.3157584,zhao-etal-2019-gender}, coreference resolution systems \cite{zhao-etal-2019-gender,rudinger-etal-2018-gender}, text classifier decisions \cite{10.1145/3287560.3287572,Jentzsch_2022}, and LLM-generated text \cite{10.5555/3157382.3157584,kotek2023genderbias,wan-etal-2023-kelly,dhingra2023queerpeoplepeoplefirst,chakrabarty2025readmeready,soundararajan-delany-2024-investigating,wu-ebling-2024-investigating,info16050358}. In turn, growing attention has focused on the behavior of LLMs in high-stakes selection contexts \cite{hickman2024performance,phillips2024hacking,henkel2024can,leong2024combining,li2024benchmarking,li2024llms,wang2024jobfair,chakrabarty2024mm,karvonen2025robustlyimprovingllmfairness}, developing resources to support safe and fair application of LLMs in such areas.

Together, these studies establish the persistence of gender cues in professional and meritocratic artifacts and the potential of NLP systems to perpetuate social bias. While bias in AI-driven evaluation can arise from multiple stages of the pipeline including the model, data, and evaluation \cite{mujtaba2024fairness}, we focus specifically on bias encoded in the language of letters of recommendation (LoRs).

Prior work on LoRs has largely examined gendered language through descriptive or correlational analyses. \cite{fu2023gender} identify gender-linked themes and stylistic patterns in biomedical LoRs using NLP and LLM-based methods, while \cite{vasan2024neurotology,vasan2024rhinology} document persistent gender differences in evaluative language in medical fellowship applications. However, these studies do not intervene on the identified cues or assess whether their removal alters downstream model behavior. In contrast, our work adopts a data-centric, interventional approach. We mask interpretable implicit gender cues in LoRs and quantify the resulting change in model performance, providing a direct assessment of how such linguistic patterns contribute to gender leakage and the extent to which data-level mitigation can reduce it.

\section{Framework}

\begin{figure*}[htbp]
    \centering
    \includegraphics[width=\linewidth]{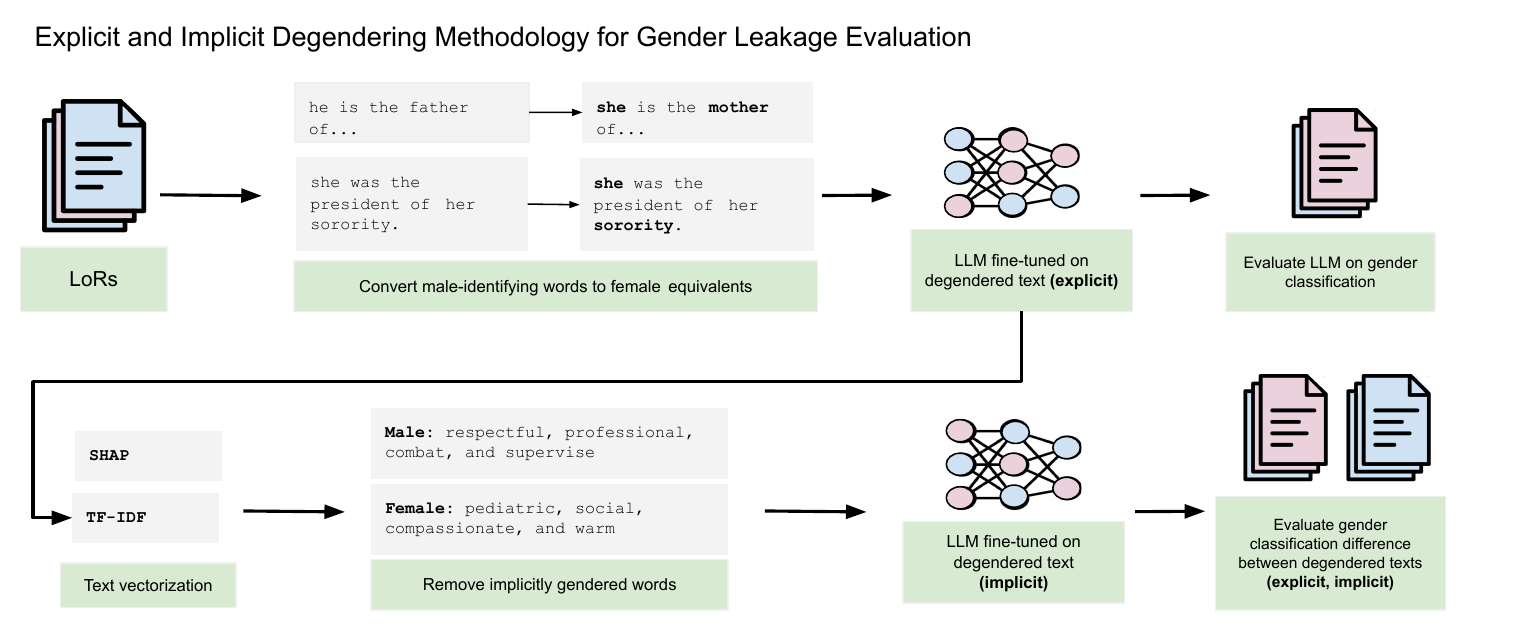}
    \caption{Overview of our case-study pipeline: corpus construction, explicit de-gendering, model training, interpretability audit, and implicit de-gendering.}
    \label{fig:workflow}
\end{figure*}

\subsection{Problem Formulation}
Let $\mathcal{D}=\{(x_i,y_i)\}_{i=1}^{N}$ be a corpus of letters of recommendation (LoRs), where $x_i \in X$ denotes the $i$-th document and $y_i\in\{0,1\}$ encodes the self-identified gender of the applicant (\textbf{0} for female, \textbf{1} for male).  
Our goal is twofold:
\begin{enumerate}[noitemsep,nolistsep]
    \item learn a mapping $f_{\theta}:X\to\{0,1\}$ that predicts the gender from $x$ as $\hat{y} = f_{\theta}(x)$, and
    \item quantify features or tokens $v \in V$ derived from $x$ using a function $\phi:V\to\mathbb{R}$ and suppress the lexical evidence that enables such prediction $\hat{y}$.
\end{enumerate}
Formally, we view $f_{\theta}$ as a text-based \emph{gender classifier} with parameters $\theta$ built atop a pre-trained Transformer encoder~\cite{vaswani2017attention} and optimized via. \emph{task-specific fine-tuning}. Also, features in our case are considered as tokens derived from the LoR texts.

\subsection{Training a Gender Classifier}

\subsubsection{Model Architecture}
Our experiments use three language model variants on the LoR dataset to build our gender classifier. Our choice of models was mostly based on the previous works done in this area which clearly establishes BERT and its variant as some of the best transformer-based models for gender classification on textual data. Moreover, out of the open-source large language models that supports text classification, Llama2 seemed to be an ideal candidate for experimentation given the resources available.

\paragraph{DistilBERT}
A six-layer student network distilled from {\sc bert-base} ($\sim$\,66\,M parameters vs.\ 110\,M) that preserves $\ge95\%$ of the original performance while being 40\% smaller and 60\% faster~\cite{sanh2020distilbertdistilledversionbert}.

\paragraph{RoBERTa}
A robustly optimized BERT derivative trained with larger mini-batches, dynamic masking, and the removal of the next-sentence objective, yielding superior downstream accuracy~\cite{liu2019robertarobustlyoptimizedbert}.

\paragraph{Llama 2}
A 7B parameter decoder-only model pre-trained on 2T tokens~\cite{touvron2023llama}, fine-tuned using parameter-efficient LoRA adapters ~\cite{hu2022lora} inserted into attention and feedforward layers. This allows efficient classification on limited GPU memory while preserving the core weights.

For all models, the final hidden state associated with the canonical \texttt{[CLS]} token ($\mathbf{h}_i \in \mathbb{R}^{d}$) is passed through a trainable affine head $\hat{y}_i\!=\!\sigma\!\bigl(\mathbf{w}^{\top}\mathbf{h}_i)$ with parameters $\mathbf{w}$ where $\sigma$ denotes the logistic function and $d$ is the dimension of the hidden state, predicting the gender category.

\subsubsection{Data Processing}
Our initial data processing pipeline includes a regex-based token matching filter, replacing all explicit gender-identifying tokens (names, titles, pronouns, and kinship terms) with special tokens or their female counterparts. The resultant filtered texts were used for training and evaluating the baseline classifier. Upon building a baseline classifier, each LoR was further subjected to an automatic \textit{de-gendering} filter $g_{\phi}$ such that $\tilde{X} = g_{\phi}(X)$, based on the the gender predictability factor $\phi$ as shown in Figure. \ref{fig:workflow}. The dataset was randomly partitioned into $\mathcal{D}_{\mathrm{train}}$:$\mathcal{D}_{\mathrm{val}}$:$\mathcal{D}_{\mathrm{test}}=80:10:10$. Tokenization follows the standard BERT based tokenization scheme with a maximum sequence length of $L{=}512$.

\subsubsection{Evaluation Metrics}
We report accuracy, precision, recall, and macro-averaged $F_{1}$ score on $\mathcal{D}_{\mathrm{test}}$. TF-IDF and SHAP values are qualitatively inspected to validate the efficacy of the de-gendering filter.

\subsection{Quantifying Implicitly Gendered Tokens}
\paragraph{SHAP} To inspect residual gender leakage, we employ SHAP \emph{(SHapley Additive exPlanations)} ~\cite{NIPS2017_8a20a862}. Formally, let $f: \mathbb{R}^n \to \mathbb{R}$ denote a predictive model, and let $x \in \mathbb{R}^n$ represent an instance with $n$ features. The goal is to decompose the output $f(x)$ as a linear combination of feature contributions such that $f(x) = \phi_0 + \sum_{i=1}^n \phi_i$, where $\phi_0$ is the base value (expected output over the dataset) and $\phi_i$ represents the marginal contribution of feature $i$ to the deviation from the base. The SHAP value $\phi_i$ for a feature $i$ is computed by taking the average marginal contribution of that feature over all possible subsets $S \subseteq \{1, \dots, n\} \setminus \{i\}$, defined as:
$\phi_i = \sum_{S \subseteq N \setminus \{i\}} \frac{|S|!(n - |S| - 1)!}{n!} \left[ f_{S \cup \{i\}}(x_{S \cup \{i\}}) - f_{S(x_S)} \right]$ where $f_S(x_S)$ denotes the model trained (or approximated) using only features in subset $S$, and $x_S$ is the projection of $x$ onto $S$. Here, $\phi_i$ denotes the SHAP value associated with token $v_i$, quantifying how much that token or word contributes to the model's deviation from the base prediction over the dataset. Positive SHAP values indicate that the word pushes the prediction toward a particular gender, while negative values push it away. For instance, high-magnitude SHAP values associated with occupational terms ("\textit{nurse}", "\textit{engineer}") reveal how the model associates language features with gender.

\paragraph{TF-IDF} By computing the TF-IDF (Term Frequency-Inverse Document Frequency) scores of words across documents labeled by gender, one can determine which terms are most characteristic or discriminative of each gender class. TF-IDF boosts terms that are frequent in a document but rare across the corpus. If certain terms consistently have higher TF-IDF scores in documents of a particular gender, then those terms clearly show a strong contributing factor influencing the prediction towards that particular gender.

\section{Experiments}
\subsection{Classification}
In our classification step, we begin by training a language model on the original LoR texts in which only the applicant names are anonymized. Since this version of the data contains explicit gender indicators such as pronouns (\textit{he}, \textit{she}) and titles (\textit{Mr.}, \textit{Mrs.}), we expect the model to predict applicant gender with near-perfect accuracy, serving as our baseline. The model used for this experiment was DistilBERT.

We then train a series of models on de-gendered versions of the text $\tilde{X}$, where all explicit gender-identifying tokens have been replaced with their female counterparts. This allows us to examine whether gender can still be inferred from more subtle linguistic patterns. For these experiments, we fine-tune transformer-based models, which includes DistilBERT, RoBERTa, and Llama 2, to evaluate their performance on the gender classification task in the absence of overt gender signals \cite{sanh2020distilbertdistilledversionbert}. 

\subsubsection{Dataset}
Our dataset consists of $8,992$ recommendation letters, each written on behalf of candidates applying to a major U.S. anesthesiology residency program. Of these, $2,787$ letters were written for female applicants and $6,205$ for male applicants, meaning approximately $31\%$ of the dataset represents female applicants and $69\%$ represents male applicants. Applicant gender was self-reported as either "male" or "female" and was contained in the letters' metadata. To preserve anonymity, applicant names in the letters were replaced with fixed special tokens such as \texttt{FIRST\_NAME}, \texttt{MIDDLE\_NAME}, \texttt{LAST\_NAME}, or \texttt{IDENTIFIER}.

\subsubsection{Data Processing Pipeline}
To de-gender the original dataset, i.e. to neutralize explicit gender-identifying tokens, we compiled a comprehensive list of gendered terms by aggregating entries from two publicly available sources: Bias-BERT and GN-GloVe \cite{Jentzsch_2022,zhao2018learninggenderneutralwordembeddings}. These lists include both obvious gender markers (e.g., \textit{he}, \textit{she}, \textit{Mr.}, \textit{Ms.}) and less immediately obvious terms with clear gender associations (e.g., \textit{husband}, \textit{father}, \textit{brother}, \textit{actor}, \textit{actress}, \textit{fraternity}, \textit{sorority}). By excluding terms such as ``\textit{father}" or ``\textit{sorority}", we ensure that gender cannot be inferred from sentences like ``\textit{he is the father of...}” or ``\textit{she was the president of her sorority.}”

To ensure comprehensive coverage, we used regular expressions to capture variations of each term, including plural forms, verb tenses, contractions, punctuation, and positioning within a sentence. This allowed us to detect and replace forms such as "\textit{she's}", "\textit{husband.}", and "\textit{mothers!}" with high precision. We then replaced any token appearing in the aggregated list with its counterpart from a single gender class to generate an \textbf{E}xplicitly \textbf{D}e-\textbf{G}endered  (\textbf{EDG}) dataset. Specifically, all explicitly male-identifying terms were converted to their female equivalents. While we considered using neutral placeholders (e.g., \textit{they}, \textit{them}), this approach risked disrupting the grammatical structure of the letters and introducing unnatural linguistic artifacts. By consistently converting all gendered terms to female, we preserved grammatical fluency while preventing the model from relying on overt gender cues, thereby encouraging it to learn from subtler, implicit gender signals embedded in the language.

\subsubsection{Training Setup}
Using the EDG dataset, we trained DistilBERT, RoBERTa, and fine-tuned Llama 2 models. These Transformer-based pre-trained models were selected due to their state-of-the-art performance on a wide range of downstream classification tasks. For each model architecture, we conducted independent hyper-parameter sweeps. DistilBERT and RoBERTa were trained using HuggingFace's Trainer API, while Llama 2 used Parameter-Efficient Fine-Tuning (PEFT) via  Low-Rank Adaption (LoRA). The parameters were optimized by Adam optimizer ($\beta_1=0.9,\beta_2=0.999,\epsilon=10^{-8}$) and a linear learning-rate schedule. Unless otherwise noted, the training runs adopt the hyper-parameters mentioned in Appendix \ref{sec:hyperparameter-opt} and were executed on Intel Xeon Gold processors and NVIDIA A100 (40\,GB) GPU hardware.

We first trained \textbf{DistilBERT} model on the original corpus (non-EDG) - retaining all gender markers - to establish an upper bound on gender learnability, using an $80:10:10$ stratified train-validation-test split, the same classification head, and only the first two encoder layers unfrozen.  An identical configuration was then applied to the \emph{de-gendered} corpus (EDG), this time keeping every sentence in each letter to preserve potential implicit gender cues.  Finally, to assess model choice on the de-gendered data, we fine-tuned \textbf{RoBERTa} and \textbf{Llama 2} on the same splits with their individually tuned best settings. Further details of the hyper-parameters optimization are provided in the Appendix \ref{sec:hyperparameter-opt}.

\subsection{Baseline Evaluation}
We evaluated our trained models on a held-out test set that was not used during training. By measuring the classification performance over multiple runs, we assess how effective our initial matching-based de-gendering method is at neutralizing explicit gender signals, and whether the model can still infer gender from implicit cues. This helps us evaluate both the strength of the remaining bias in the text and the robustness of the model's predictions. To assess performance, we considered standard classification metrics, including accuracy, precision, recall, and $F_1$ score. Given the class imbalance in our dataset—with a larger proportion of letters written for male applicants-we placed particular emphasis on the macro $F_1$ score, which provides a more balanced evaluation by averaging the $F_1$ scores across both classes independently. This ensures that the model's performance on the minority class (female applicants) is not overshadowed by the majority class.

\subsection{Selecting Implicitly Gendered Tokens}
To interpret our classification results, we explored the linguistic artifacts and implicit signals within the data that the model may be using to identify gender, offering insight into the subtler ways in which gendered language cues can manifest in recommendation letters. 

\begin{figure*}[t]
  \centering
  \begin{subfigure}[b]{0.3\textwidth}
    \centering
    \includegraphics[width=\textwidth]{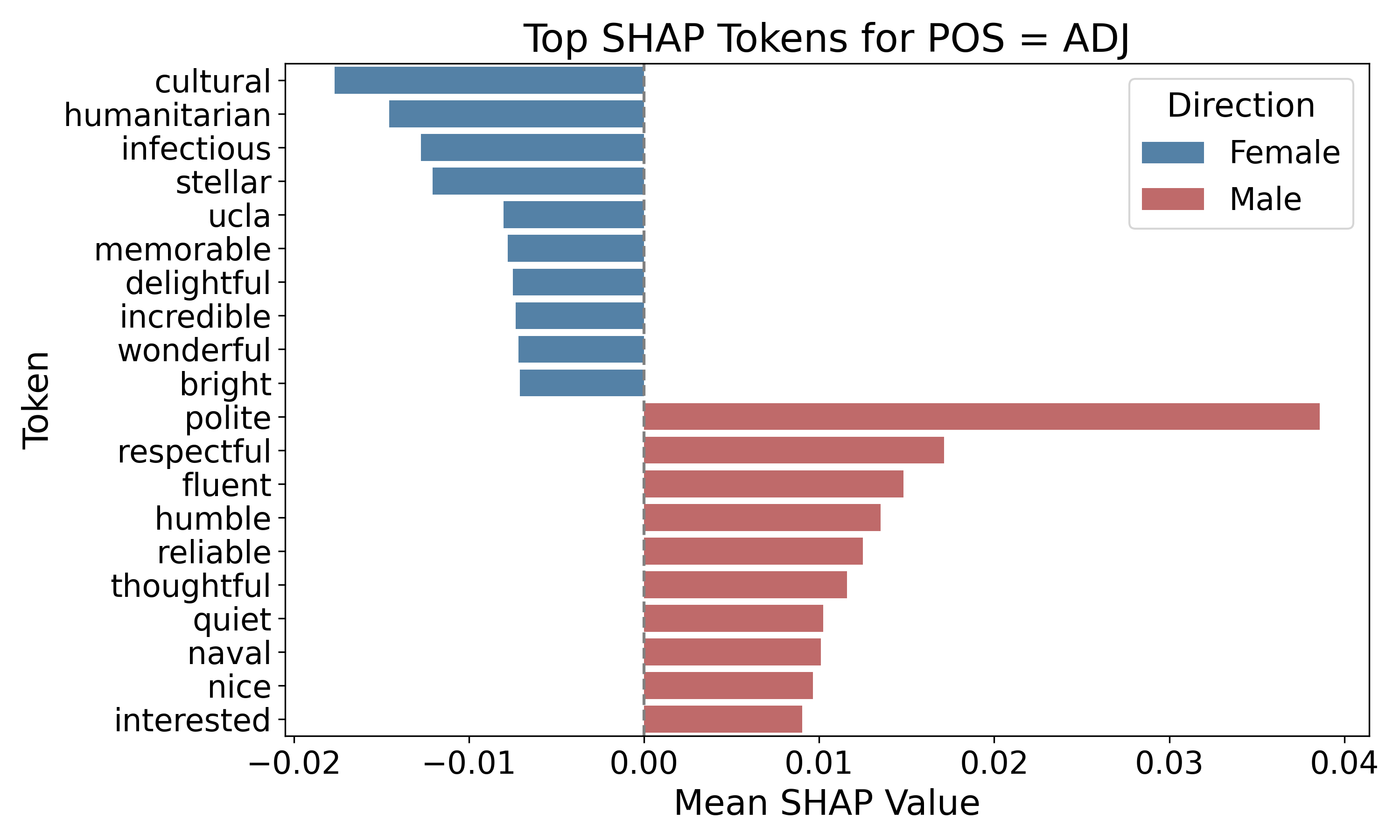}
    \caption{Adjectives (DistilBERT)}
    \label{fig:distil_shap_values_adjectives}
  \end{subfigure}
  \hfill
  \begin{subfigure}[b]{0.3\textwidth}
    \centering
    \includegraphics[width=\textwidth]{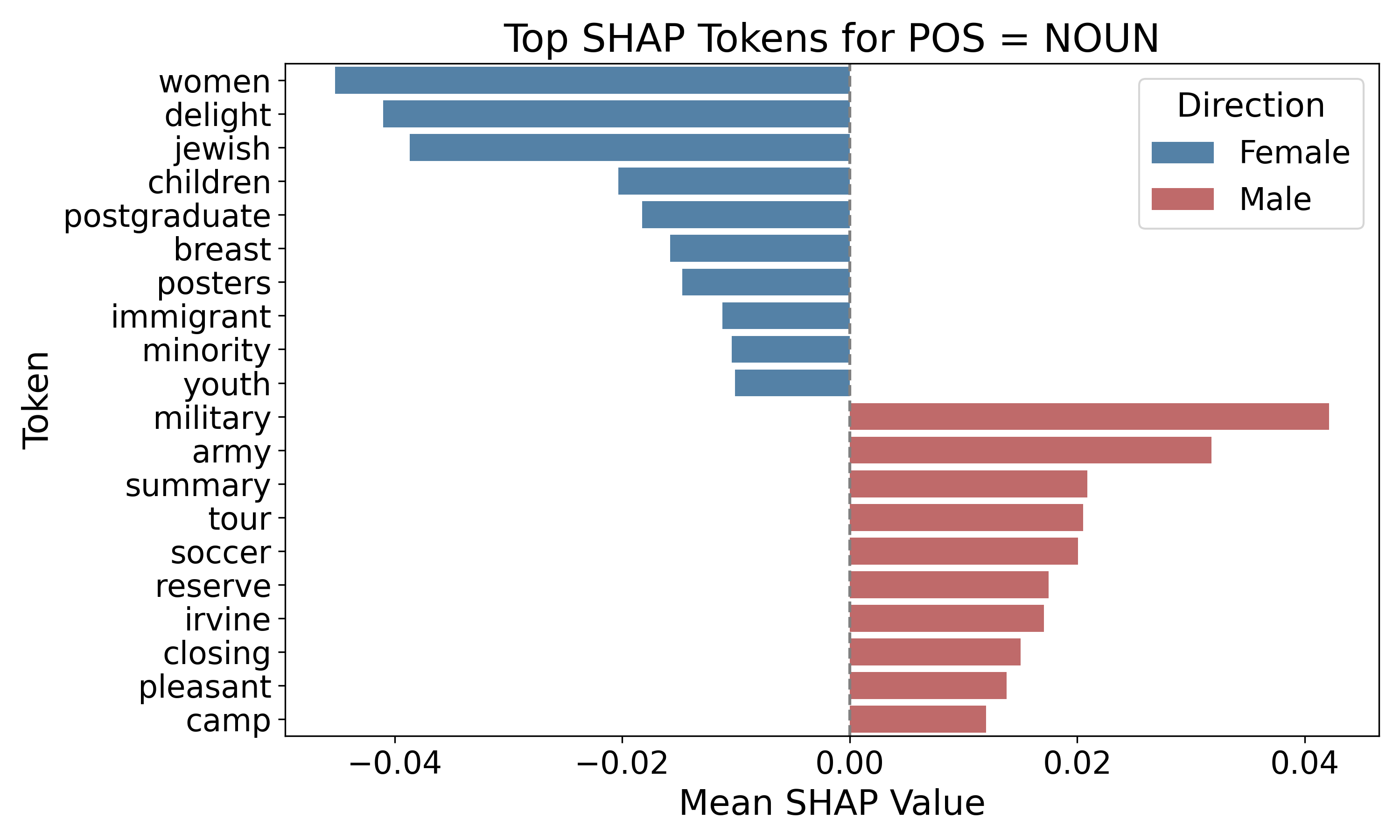}
    \caption{Nouns (DistilBERT)}
    \label{fig:distil_shap_values_nouns}
  \end{subfigure}
  \hfill
  \begin{subfigure}[b]{0.3\textwidth}
    \centering
    \includegraphics[width=\textwidth]{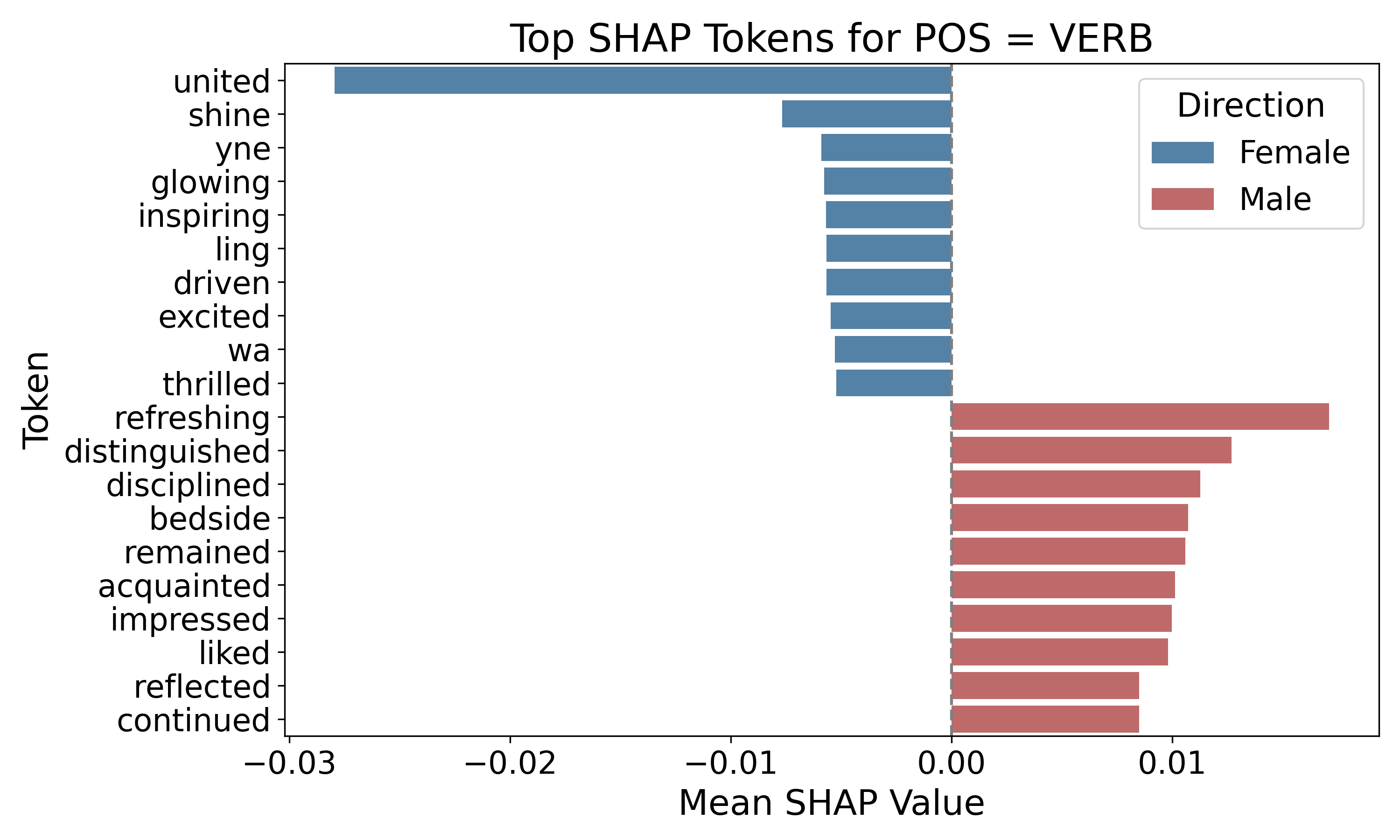}
    \caption{Verbs (DistilBERT)}
    \label{fig:distil_shap_values_verbs}
  \end{subfigure}

  \centering
  \begin{subfigure}[b]{0.3\textwidth}
    \centering
    \includegraphics[width=\textwidth]{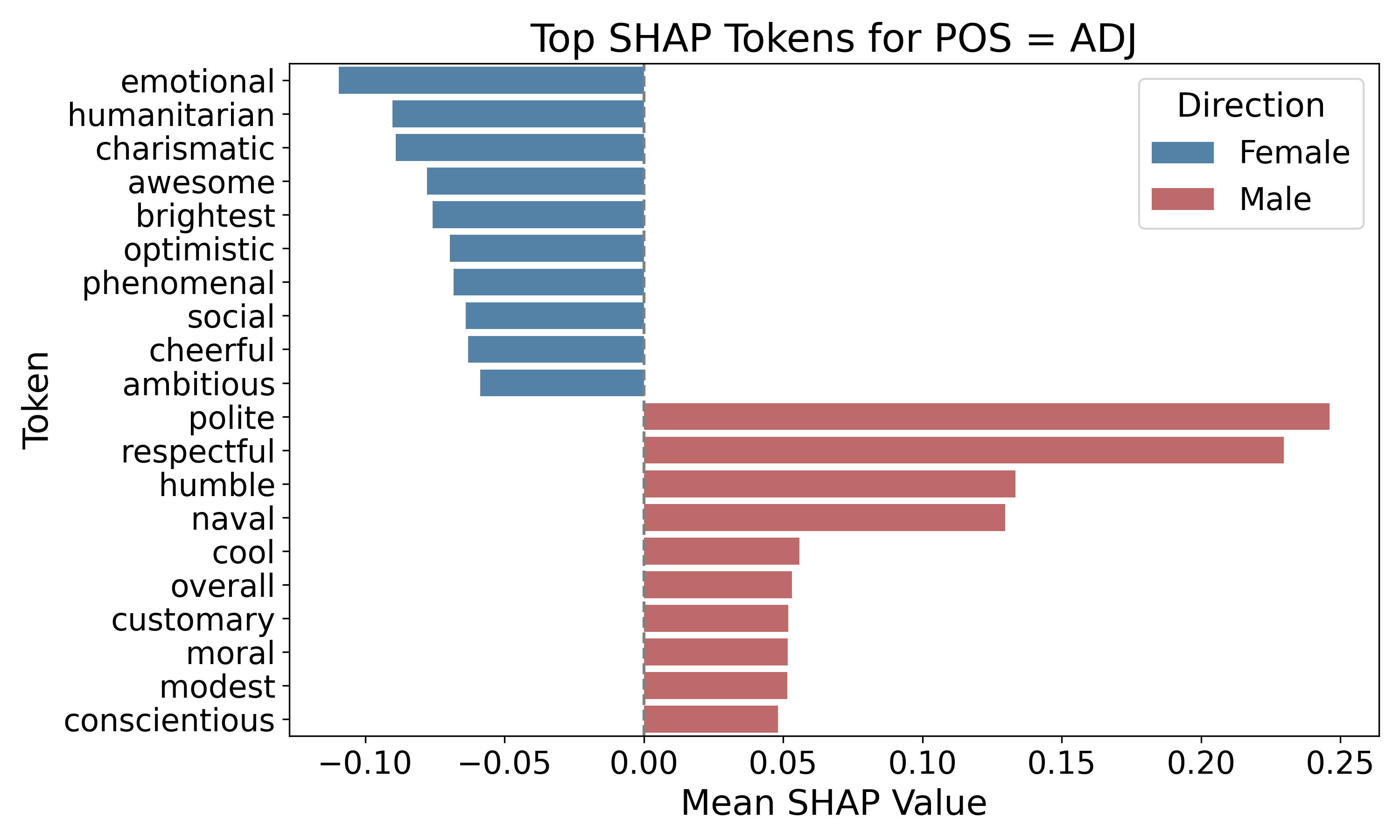}
    \caption{Adjectives (RoBERTa)}
    \label{fig:roberta_shap_values_adjectives}
  \end{subfigure}
  \hfill
  \begin{subfigure}[b]{0.3\textwidth}
    \centering
    \includegraphics[width=\textwidth]{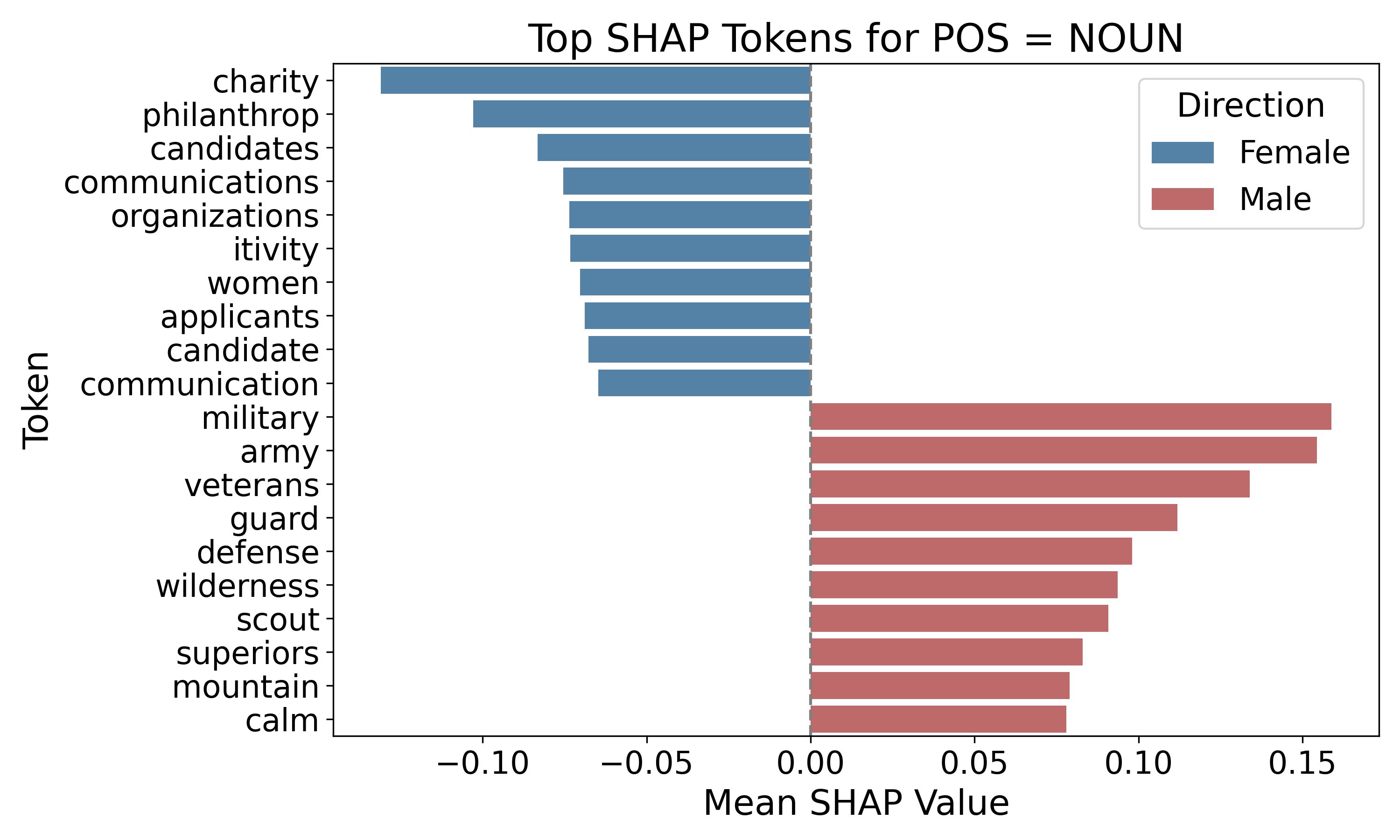}
    \caption{Nouns (RoBERTa)}
    \label{fig:roberta_shap_values_nouns}
  \end{subfigure}
  \hfill
  \begin{subfigure}[b]{0.3\textwidth}
    \centering
    \includegraphics[width=\textwidth]{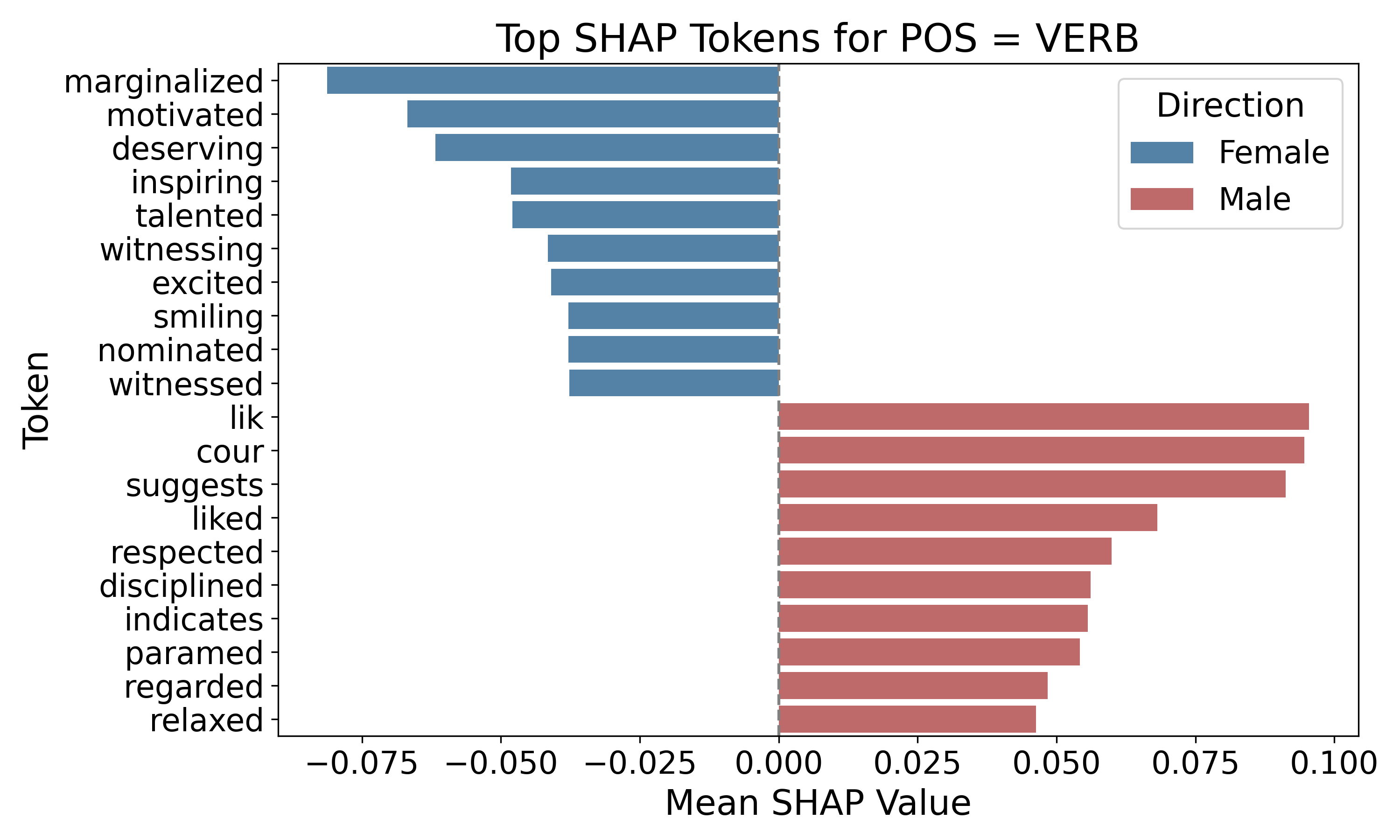}
    \caption{Verbs (RoBERTa)}
    \label{fig:roberta_shap_values_verbs}
  \end{subfigure}

  \centering
  \begin{subfigure}[b]{0.3\textwidth}
    \centering
    \includegraphics[width=\textwidth]{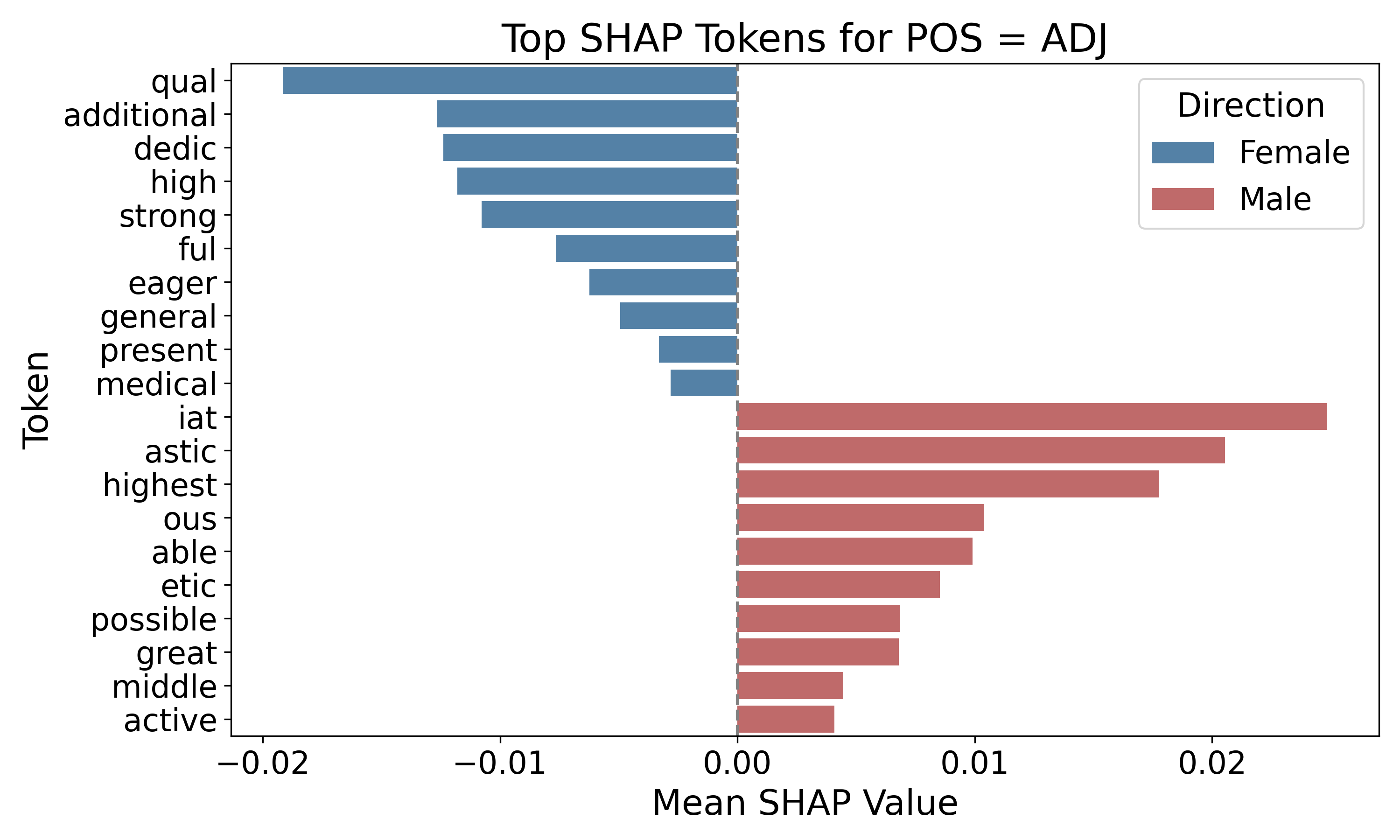}
    \caption{Adjectives (Llama 2)}
    \label{fig:llama_shap_values_adjectives}
  \end{subfigure}
  \hfill
  \begin{subfigure}[b]{0.3\textwidth}
    \centering
    \includegraphics[width=\textwidth]{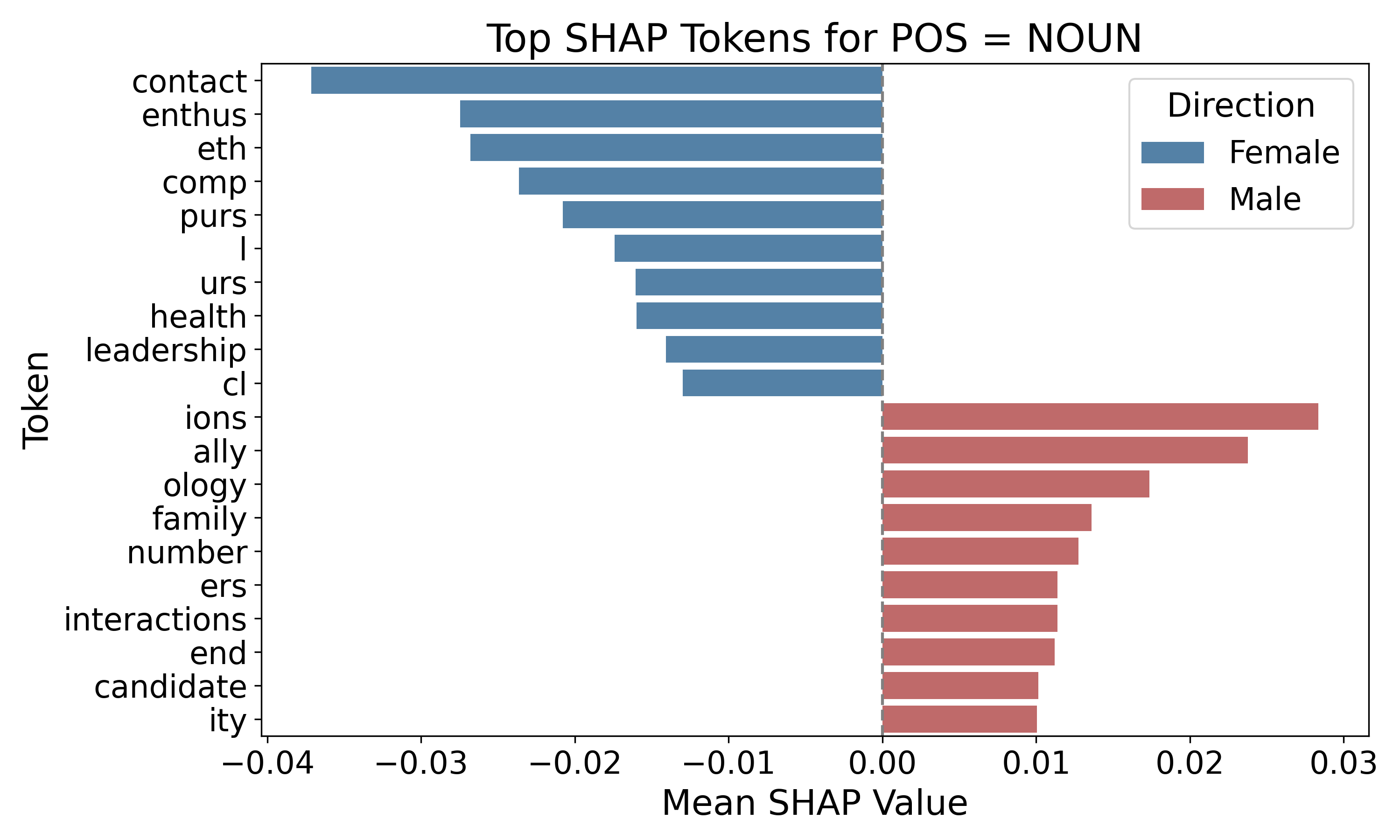}
    \caption{Nouns (Llama 2)}
    \label{fig:llama_shap_values_nouns}
  \end{subfigure}
  \hfill
  \begin{subfigure}[b]{0.3\textwidth}
    \centering
    \includegraphics[width=\textwidth]{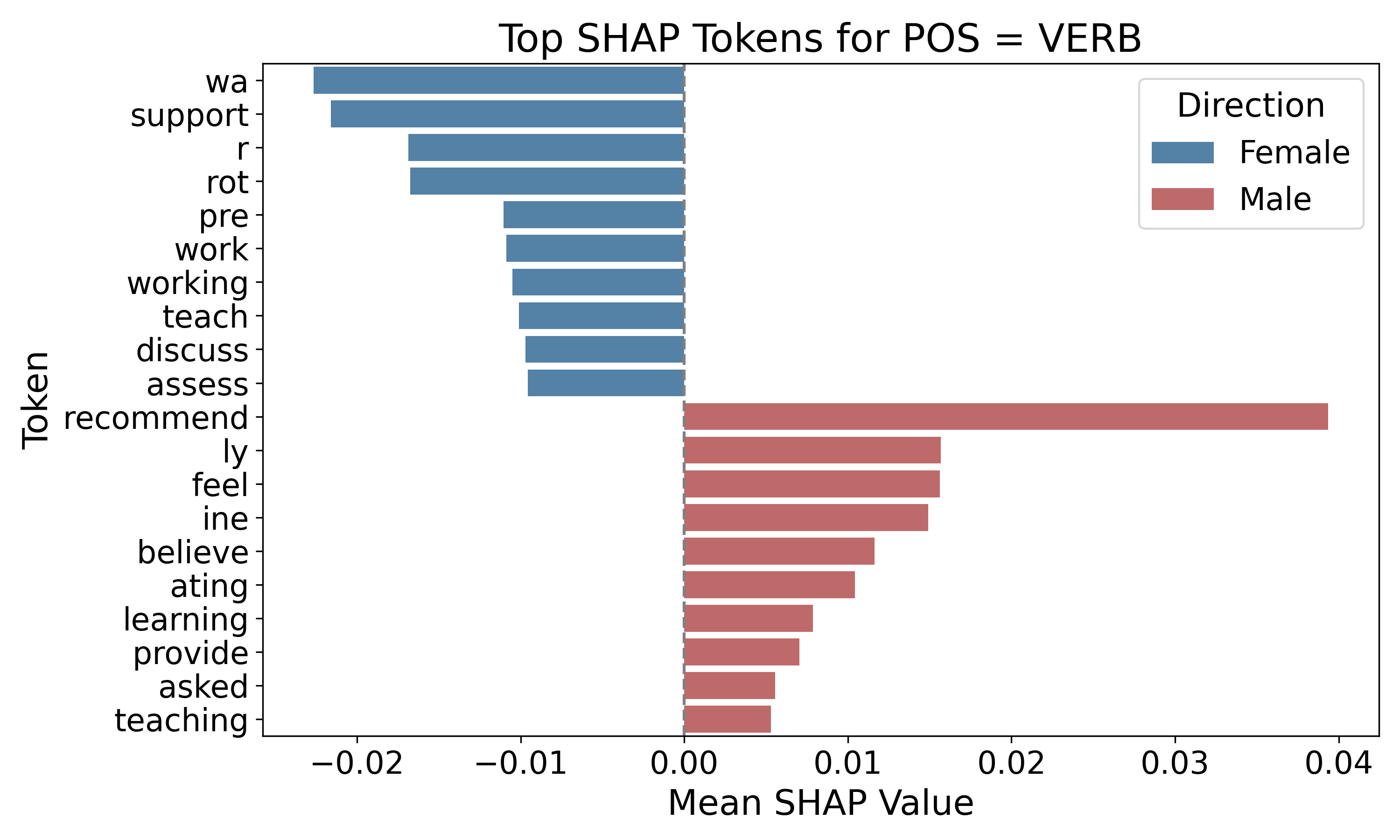}
    \caption{Verbs (Llama 2)}
    \label{fig:llama_shap_values_verbs}
  \end{subfigure}
  \caption{Top 10 SHAP tokens with their corresponding values(+/-) for both genders (male/ female) grouped by part-of-speech category, identified by DistilBERT(last two encoder layers unfrozen), RoBERTa and Llama 2}
  \label{fig:shap_values_pos_combined}
\end{figure*}

\subsubsection{SHAP Analysis}
To better understand the sources of gender leakage within the de-gendered text, we used SHAP \cite{NIPS2017_8a20a862} values to interpret the model's predictions. SHAP assigns importance scores to individual tokens, allowing us to identify which words contributed most to the model's classification of an applicant as male or female. These influential tokens offer insight into the subtle linguistic cues the model relies on in the absence of explicit gender indicators. SHAP analysis was applied to all three of the models trained on our de-gendered dataset. For DistilBERT and RoBERTa, SHAP values were computed across the full set of recommendation letters. For Llama, we used a random sample of 100 letters due to the higher computational cost. We then extracted the tokens with the highest and lowest mean SHAP values across letters-associated with male and female predictions, respectively. These were grouped by part of speech to aid in our interpretation. To reduce noise, we only considered tokens that appeared at least 20 times across the dataset.

\subsubsection{Interpretation using TF-IDF} 
In addition to model-based interpretation, we conducted a complementary analysis using TF-IDF. Rather than relying on model outputs, we applied the TF-IDF algorithm directly to the EDG dataset to identify terms that are most distinctive to male and female applicant letters, respectively. This approach provides a model-agnostic view of potential gendered patterns in the language used. We first aggregated all de-gendered recommendation letters for female applicants into a single document, and all de-gendered letters for male applicants into another. We then applied the TF-IDF algorithm to these two documents, computing a score for each token based on its frequency and uniqueness within its respective group. Tokens with higher TF-IDF scores are those that appear frequently and are particularly distinctive to one group, offering potential insight into the subtle linguistic patterns that may encode gender even after explicit identifiers have been neutralized. More details are provided in Appendix \ref{sec:tf-idf_analysis}.

\subsection{Effect of Implicit Gender Tokens}

\begin{table}[t]
\caption{Token-level gender prediction flips for TF-IDF derived tokens: comparison of top 10 tokens whose removal most frequently causes DistilBERT model prediction to flip from female to male or vice versa.}

\centering
\resizebox{0.7\columnwidth}{!}{%
\begin{tabular}{lccc}
\hline
\textbf{Token} & \textbf{F $\rightarrow$ M Count} & \textbf{M $\rightarrow$ F Count} & \textbf{Absolute Difference} \\
\hline
support & 12 & 23.56 & 11.56 \\
research & 14 & 24.64 & 10.64 \\
number & 4 & 11.98 & 7.98 \\
anesthesia & 13 & 20.96 & 7.96 \\
rotation & 26 & 18.26 & 7.74 \\
medicine & 17 & 23.29 & 6.29 \\
believe & 8 & 14.2 & 6.2 \\
leadership & 3 & 9.02 & 6.02 \\
professional & 6 & 11.75 & 5.75 \\
year & 18 & 23.3 & 5.3 \\
\hline
\end{tabular}%
}
\label{tab:tfidf-token-flip-analysis}
\end{table}

\begin{table}[t]
\caption{Token-level gender prediction flips for SHAP derived tokens: comparison of top tokens whose removal most frequently causes DistilBERT model prediction to flip from female to male or vice versa.}

\centering
\resizebox{0.7\columnwidth}{!}{%
\begin{tabular}{lccc}
\hline
\textbf{Token} & \textbf{F $\rightarrow$ M Count} & \textbf{M $\rightarrow$ F Count} & \textbf{Absolute Difference} \\
\hline
liked & 6 & 0.38 & 5.62 \\
impressed & 14 & 8.93 & 5.07 \\
bedside & 6 & 2.8 & 3.2 \\
stellar & 3 & 0.37 & 2.63 \\
summary & 1 & 3.16 & 2.16 \\
thoughtful & 4 & 1.97 & 2.03 \\
acquainted & 2 & 0 & 2.00 \\
women & 2 & 0 & 2.00 \\
reliable & 1 & 2.78 & 1.78 \\
united & 2 & 0.38 & 1.62 \\
\hline
\end{tabular}%
}
\label{tab:shap-token-flip-analysis}
\end{table}

After identifying implicitly gendered tokens using the TF-IDF and SHAP methods described in the previous sections, we selected the top 10 tokens for each part of speech (noun, verb, adjective), identified separately for each gender and for each method independently. This process produced two distinct datasets: one based on TF-IDF-selected tokens and another based on SHAP-selected tokens. In both cases, the selected tokens were altered in texts that had already been stripped of explicit gender-identifying terms. For DistilBERT and RoBERTa, this involved substituting the tokens with the model's masked token; for Llama 2, which does not support masking, we used the unknown token instead. This intervention was applied consistently across both token sets, allowing us to evaluate the extent to which implicitly gendered language contributed to model performance, even in the absence of explicit gender cues. We then re-trained all three models on each of the new datasets (EDG w/o SHAP tokens and EDG w/o TF-IDF tokens).

\subsubsection{Analyzing Prediction Flips from Token Removal}
A deeper analysis of implicitly gendered tokens was conducted by focusing on a subset of recommendation letters that met two criteria: (1) they were correctly classified by the model trained on letters with only explicit gender-identifying tokens replaced (the EDG dataset), and (2) they were misclassified by the model trained on letters with both explicit and implicit tokens replaced.
For this analysis, we used only the DistilBERT model due to its strong performance and computational efficiency. This allowed us to examine how implicitly biased tokens influenced model predictions when neutralized. For each such token, we counted how often replacing it with the model's masked token (or unknown token in the case of Llama) caused the model's prediction to flip from female to male or from male to female. A flip in prediction direction suggests that the replaced token may have carried meaningful gender-associated information that influenced the model's original prediction.

To mitigate the effects of class imbalance, we performed random sub-sampling of the majority class within this subset, repeated the token-flip counting process across multiple runs, and averaged the results. This approach provided a more balanced view of the influence of individual tokens. Table~\ref{tab:tfidf-token-flip-analysis} \& \ref{tab:shap-token-flip-analysis} presents the TF-IDF and SHAP-derived tokens along with their corresponding flip counts respectively. 

\section{Results}
\begin{table*}[htbp]
\caption{Comparison of DistilBERT, RoBERTa, and Llama 2 performance on test EDG(Explicitly De-Gendered) dataset including individual class metrics (Precision, Recall and $F_1$) and aggregate metrics(Macro + Weighted Precision, Recall and $F_1$).}

\centering
\renewcommand{\arraystretch}{1.5}
\setlength{\tabcolsep}{4pt}
\resizebox{\textwidth}{!}{%
\begin{tabular}{lcccccccccccccc}
\toprule
\textbf{Model} & \textbf{Gender} & \textbf{Precision} & \textbf{Recall} & \textbf{$F_1$} &
\textbf{Acc.} & \textbf{Macro Precision} & \textbf{Macro Recall} & \textbf{Macro $F_1$} &
\textbf{Wtd. Precision} & \textbf{Wtd. Recall} & \textbf{Wtd. $F_1$}\\
\midrule
\multirow{ 2}{*}{DistilBERT} & Female & 0.413 & 0.403 & 0.41 & \multirow{ 2}{*}{0.633} & \multirow{ 2}{*}{0.573} & \multirow{ 2}{*}{0.573} & \multirow{ 2}{*}{0.577} & \multirow{ 2}{*}{0.637} & \multirow{ 2}{*}{0.633} & \multirow{ 2}{*}{0.633} \\
& Male & 0.733 & 0.74 & 0.737 & & & & & & & \\\hline
\multirow{ 2}{*}{RoBERTa} & Female & 0.39 & 0.397 & 0.387 & \multirow{ 2}{*}{0.61} & \multirow{ 2}{*}{0.55} & \multirow{ 2}{*}{0.55} & \multirow{ 2}{*}{0.553} & \multirow{ 2}{*}{0.617} & \multirow{ 2}{*}{0.61} & \multirow{ 2}{*}{0.61} \\
& Male & 0.72 & 0.703 & 0.68 & & & & & & & \\\hline
\multirow{ 2}{*}{Llama 2} & Female & 0.395 & 0.444 & 0.418 & \multirow{ 2}{*}{0.612} & \multirow{ 2}{*}{0.560} & \multirow{ 2}{*}{0.563} & \multirow{ 2}{*}{0.561} & \multirow{ 2}{*}{0.623} & \multirow{ 2}{*}{0.612} & \multirow{ 2}{*}{0.615} \\
& Male & 0.725 & 0.682 & 0.703 & & & & & & & \\
\bottomrule
\end{tabular}
}
\renewcommand{\arraystretch}{1}
\label{tab:full-model-comparison}
\end{table*}

From Table~\ref{tab:classification-comparison-all-expanded}, we see that when explicit gender markers were preserved, the DistilBERT baseline achieved an almost perfect macro $F_1 > 0.95$, confirming that surface cues such as "\textit{he}" or professional titles like "\textit{Ms.}" virtually guarantee correct gender classification. After converting all explicit gender-identifying tokens to their female equivalents (the EDG dataset) and retraining, the same model architecture still obtained a macro $F_1$ score of $0.577$ with an overall accuracy of $0.633$ on a held‑out test set, indicating that subtler linguistic patterns continue to signal applicant gender.  Alternative models performed worse: RoBERTa reached a macro $F_1$ score of $0.553$, and the considerably larger Llama 2 yielded a comparable $F_1$ score of $0.561$ while incurring far greater computational cost. 

The classification results of re-evaluation with their comparison to the model trained on the text with only the explicit gender identifying tokens replaced, are also shown in Table~\ref{tab:classification-comparison-all-expanded}. For the DistilBERT model, replacing the implicitly gendered tokens identified via SHAP and TF-IDF with a masked token leads to a drop in macro $F_1$ score by $\sim 2.7\%$ and $\sim 1.4$\% respectively. This decline highlights the extent to which these tokens contributed to the model's ability to predict applicant gender in the absence of explicit gender-identifying terms.

\begin{table*}[t]

\caption{Comparison of DistilBERT, RoBERTa, and Llama 2 performance across three datasets: EDG (Explicitly De-Gendered), EDG with top SHAP-identified gender tokens removed, and EDG with top TF-IDF-identified gender tokens removed showing improvements(lowering) in macro precision, recall, and $F_1$ scores}

\centering
\renewcommand{\arraystretch}{1.5}
\setlength{\tabcolsep}{4pt}
\resizebox{\textwidth}{!}{%
\begin{tabular}{l|cccc|cccc|cccc}
\hline
\multirow{2}{*}{\textbf{Dataset}} 
& \multicolumn{4}{c|}{\textbf{DistilBERT}} 
& \multicolumn{4}{c|}{\textbf{RoBERTa}} 
& \multicolumn{4}{c}{\textbf{Llama 2}} \\
\cline{2-13}
 & Acc. & Macro P & Macro R & Macro $F_1$ 
 & Acc. & Macro P & Macro R & Macro $F_1$ 
 & Acc. & Macro P & Macro R & Macro $F_1$ \\
\hline
Original (non-EDG) 
& 0.999 & 1.000 & 0.996 & 0.999 
& -- & -- & -- & -- 
& -- & -- & -- & -- \\
EDG (baseline) 
& 0.633 & 0.573 & 0.573 & 0.577 
& 0.61 & 0.55 & 0.55 & 0.553 
& 0.612 & 0.56 & 0.563 & 0.561 \\
EDG w/o SHAP Tokens 
& 0.597 & 0.553 & 0.56 & 0.55 ($\downarrow$ 2.7\%) 
& 0.585 & 0.56 & 0.57 & 0.55 $\downarrow$ 0.3\%) 
& 0.690 & 0.345 & 0.500 & 0.408 ($\downarrow$ 15.3\%) \\
EDG w/o TF-IDF Tokens 
& 0.61 & 0.567 & 0.57 & 0.563 ($\downarrow$ 1.4\%) 
& 0.555 & 0.54 & 0.54 & 0.53 ($\downarrow$ 2.3\%) 
& 0.685 & 0.344 & 0.497 & 0.407 ($\downarrow$ 15.4\%) \\
\hline
\end{tabular}
}
\renewcommand{\arraystretch}{1}
\label{tab:classification-comparison-all-expanded}
\end{table*}

\subsection{Replacing Explicit Gender Identifying Tokens}

When we replace all explicit gender-identifying tokens in the recommendation letters with their female equivalents, the performance of our baseline classifier drops significantly. This confirms that our de-gendering process is effective and substantially limits the model's ability to infer applicant gender from overt cues. However, the fact that the macro $F_1$ score remains above random chance suggests that the model trained on de-gendered text can still identify subtle, implicit gender signals. In other words, even after neutralizing explicit gendered language, the way writers describe male and female applicants still carries implicit patterns that the model can detect.

\subsection{Implicit Gender Signals}

Using our SHAP and TF-IDF analyses, we identify tokens that may carry implicit gender signals within the text. The top 10 SHAP values of adjectives, nouns and verbs for male and female candidates are shown in  \autoref{fig:distil_shap_values_adjectives}, \autoref{fig:distil_shap_values_nouns}, and \autoref{fig:distil_shap_values_verbs} respectively. Tokens more commonly associated with female recommendation letters include words like ``\textit{humanitarian}", ``\textit{delightful}", ``\textit{wonderful}", and ``\textit{children}". In contrast, tokens more frequently linked to male recommendation letters include ``\textit{respectful}", ``\textit{military}", ``\textit{combat}", and ``\textit{humble}". These patterns suggest subtle differences in how male and female applicants are described, even after explicit gender markers have been obscured.

\subsection{Replacing Implicit Gender Identifying Tokens}

Across all three models, replacing the implicitly gendered tokens identified by SHAP and TF-IDF with the corresponding model's masked token (or the unknown token in the case of Llama 2) resulted in drops in macro $F_1$ scores. This decline indicates that, even in the absence of explicit gender identifiers, the models trained on the EDG dataset relied heavily on implicit gender cues to make their predictions.

A closer analysis of the tokens whose replacement caused prediction flips further illustrates this point. For example, the token ``\textit{leadership}", when replaced, caused the model to flip its prediction from male to female $9$ times, compared to just $3$ flips in the opposite direction. This suggests that the presence of ``\textit{leadership}" strongly contributes to the model associating the letter with a male applicant. Similarly, tokens like ``\textit{rotation}" and ``\textit{liked}" more often flipped predictions from female to male when replaced, indicating that their presence is more commonly associated with female applicants.

\section{Summary}\label{sec:conclusion}
In this case study, we investigated the presence of gendered language in academic LoRs from a medical residency program. Despite the replacement of explicit gender identifiers, such as names and gendered pronouns, with their female equivalents, our results demonstrated that encoder models as well as LLMs could achieve above-chance accuracy in applicant gender classification. The models achieved 63\% accuracy (57\% macro~$F_{1}$) in predicting applicant gender which is well above chance, thus indicating substantial implicit cues. Our TF-IDF and SHAP analysis showed that specific adjectives, verbs, and nouns served as implicit indicators of gender and contributed heavily to this performance, and classification performance dropped sharply when replacing them with the model's masked token (or the \texttt{unknown} token in the case of Llama 2). After this iterative token removal process, accuracy drops to 60\%, yet still exceeds random guessing, underscoring the tenacity of gender-specific language patterns even in ostensibly neutral text.

\section{Discussion}

These findings raise important questions about the role of AI in professional evaluation and recruitment. \cite{dastinInsightAmazonScraps2018} highlights how the na{\"i}ve strategy of erasing explicit cues of gender is insufficient to prevent bias in AI recruitment solutions, especially where training data already reflects historical bias in hiring. While \cite{dastinInsightAmazonScraps2018} focuses on the entrenchment of explicit gender cues in applicant qualifications (e.g., names of all-women's colleges) and the challenge of neutralizing them, our work broadens this concern by revealing persistent AI-recognizable implicit cues of gender in LoRs. Even after neutralizing implicit cues of gender, our LLM-based gender classifiers achieve above-chance accuracy, suggesting that gender signals remain.

Moreover, recruitment methods that attempt to neutralize \textit{any} cues (explicit or implicit) for applicant attributes like gender trivialize candidate identities as attributes that can simply be switched off. They can prevent the consideration of important information about the experiences of candidates from marginalized backgrounds, and ultimately may serve to bypass or outsource diversity, equity, and inclusion efforts that should instead occur within organizations \cite{drage2022does,tilmes2022disability}. 
While neutralizing these cues could prevent gender identification from LoRs, this would inevitably also mask positive qualities and important experiences of candidates. 
Consequently, a purely technical approach to addressing bias in the evaluation and selection of candidates may be an impossible goal. 

As such, we believe that the role of AI in hiring processes must be carefully weighed. De-gendering strategies like those explored in this work may be one component of fairer AI-supported evaluation of LoRs, e.g., as a flag to reviewers to be aware of the amount of gendered language in an LoR. However, it essential to pair this with increased investment in diverse human evaluators and institutional change, such as implicit bias training.

\section{Limitations}

Here, in addition to the previous section's discussion, we acknowledge some technical, data-related, and conceptual limitations to our approach. First, with respect to neutralizing implicit gender cues, the semantic integrity and evaluative content of letters may not be preserved. Our approach replaces nouns, verbs, and adjectives with masked (or unknown) tokens; thus the resulting letters may contain incomplete and ungrammatical sentences. As such, this approach would be more appropriate for machine evaluation of letters, as human evaluators may have difficulties understanding some parts of the letters. 

Second, we found that SHAP occasionally highlights words that are split into sub-tokens during tokenization. Although SHAP applies heuristics to collapse these sub-tokens back into full words for easier interpretation, this collapsing does not always align precisely with how the model processes inputs internally. This misalignment can lead to cases where attribution scores are assigned to fragments rather than complete tokens, introducing some ambiguity into the interpretability analysis. We note this as a limitation of our approach.

Third, we note that we did not have access to applicants' full files, nor data on the outcome of residency selection decisions. As such, our work cannot explore links between gendered language and outcomes; it is instead a case study of technical de-gendering processes and the persistence of gender-identifying text as an input into downstream decision-making.  

\section*{Ethical Considerations Statement}

This work analyzes academic letters of recommendation, which contain identifiable and possibly sensitive human subjects data. As such, the collection of these data was IRB approved, and all applicant names were anonymized before analysis. All research was performed in accordance with ACM's Publications Policy on Research Involving Human Participants and Subjects.

\bibliographystyle{unsrt} 
\bibliography{references}

\appendix

\section{Hyper-parameter Optimization}
\label{sec:hyperparameter-opt}
To maximize classification performance on de-gendered LoR datasets, we conducted a limited grid search over batch size $\in \{4, 16, 32\}$, learning rate $\in [10^{-5},  3.7\times 10^{-5}]$, and weight decay $\in \{0.0, 0.03\}$ for both DistilBERT and RoBERTa. For Llama 2, we additionally tuned LoRA parameters ($r$, $\alpha$, dropout). Validation performance was monitored using macro-averaged $F_1$. The resulting best settings are reported in Tables \ref{tab:best-distilbert-hparams}, \ref{tab:best-roberta-hparams} and \ref{tab:best-llama2-hparams}.

\begin{table}[ht]
\centering
    \caption{Best hyper-parameters selected for DistilBERT and RoBERTa text classifiers..}

\begin{subtable}{0.45\textwidth}
        \centering
        \begin{tabular}{ll}
            \toprule
            \textbf{Parameter} & \textbf{Value} \\
            \midrule
            Batch size (train / eval) & 32 / 32 \\
            Epochs                    & 10 \\
            Learning rate             & $3.7 \times 10^{-5}$ \\
            Weight decay              & 0.03 \\
            \bottomrule
        \end{tabular}
        \caption{Best DistilBERT fine-tuning hyper-parameters}
        \label{tab:best-distilbert-hparams}
    \end{subtable}%
    \hfill
    \begin{subtable}{0.45\textwidth}
        \centering
        \begin{tabular}{ll}
            \toprule
            \textbf{Parameter} & \textbf{Value} \\
            \midrule
            Batch size (train / eval) & 16 / 16 \\
            Epochs                    & 6 \\
            Learning rate             & $2.0 \times 10^{-5}$ \\
            Weight decay              & 0.0 \\
            \bottomrule
        \end{tabular}
        \caption{Best RoBERTa fine-tuning hyper-parameters}
        \label{tab:best-roberta-hparams}
    \end{subtable}
    \label{tab:combined_tables}
\end{table}

\begin{table}[!htbp]

\caption{Best Llama 2 fine-tuning hyper-parameters}

\centering
\begin{tabular}{ll}
    \toprule
    \textbf{Parameter} & \textbf{Value} \\
    \midrule
    Batch size (train / eval) & 4 / 4 \\
    Epochs                    & 5 \\
    Learning rate             & $3.0 \times 10^{-5}$ \\
    Weight decay              & 0.02 \\
    LoRA rank ($r$)           & 16 \\
    LoRA alpha               & 48 \\
    LoRA dropout            & 0.15 \\
    \bottomrule
\end{tabular}
\label{tab:best-llama2-hparams}
\end{table}

\section{Topic Modeling}

\subsection{Explainability via Topic Modeling}

In addition to token-level interpretability methods, topic modeling techniques were explored as a way to uncover higher-level thematic patterns that may reflect gendered language in recommendation letters. Using these approaches, we identified recurring topics across the corpus and analyzed which ones were most predictive of each gender. This allowed us to examine broader narrative trends and associations, offering a complementary perspective to the more granular insights provided by token-level analysis.

\subsection{BERTopic}

One of the popular topic modeling techniques called \textit{BERTopic} was used to identify topics across our recommendation letters \cite{grootendorst2022bertopicneuraltopicmodeling}. Each letter was first split into individual sentences, which were then embedded and clustered based on semantic similarity. In total, \textit{BERTopic} extracted 251 distinct topics, capturing a broad range of recurring themes. To create a topic-level representation for each letter, we mapped the identified topics of individual sentences back to their originating letter. This resulted in a binary topic vector for each letter, where each entry indicates the presence (1) or absence (0) of a specific topic.

\subsection{Topic-Only Classification}

For classification, we created train and test splits using the topic vectors for each letter and trained a random forest classifier to predict gender based on these topic representations. Model performance was evaluated on the held-out test set, with results summarized in Table \ref{tab:topic-and-bert-performance}.

\begin{table*}[htbp]

\caption{Classification performance of Random Forest models using (1) only topic vectors and (2) a combination of topic vectors and DistilBERT embeddings.}

\centering
\resizebox{\textwidth}{!}{%
\begin{tabular}{lcccccccccccccc}
\toprule
\textbf{Model} & \textbf{Gender} & \textbf{Precision} & \textbf{Recall} & \textbf{$F_1$} &
\textbf{Acc.} & \textbf{Macro Precision} & \textbf{Macro Recall} & \textbf{Macro $F_1$} &
\textbf{Wtd. Precision} & \textbf{Wtd. Recall} & \textbf{Wtd. $F_1$}\\
\midrule
\multirow{2}{*}{Topic Vectors Only} & Female & 0.370 & 0.480 & 0.420 & \multirow{2}{*}{0.580} & \multirow{2}{*}{0.550} & \multirow{2}{*}{0.560} & \multirow{2}{*}{0.550} & \multirow{2}{*}{0.620} & \multirow{2}{*}{0.580} & \multirow{2}{*}{0.600} \\
& Male & 0.730 & 0.630 & 0.680 & & & & & & & \\
\midrule
\multirow{2}{*}{Topic + DistilBERT Embeddings} & Female & 0.410 & 0.440 & 0.430 & \multirow{2}{*}{0.630} & \multirow{2}{*}{0.580} & \multirow{2}{*}{0.580} & \multirow{2}{*}{0.580} & \multirow{2}{*}{0.640} & \multirow{2}{*}{0.630} & \multirow{2}{*}{0.640} \\
& Male & 0.740 & 0.720 & 0.730 & & & & & & & \\
\bottomrule
\end{tabular}
}
\label{tab:topic-and-bert-performance}
\end{table*}

\begin{figure}[htbp]
  \centering
  \includegraphics[width=\linewidth, keepaspectratio]{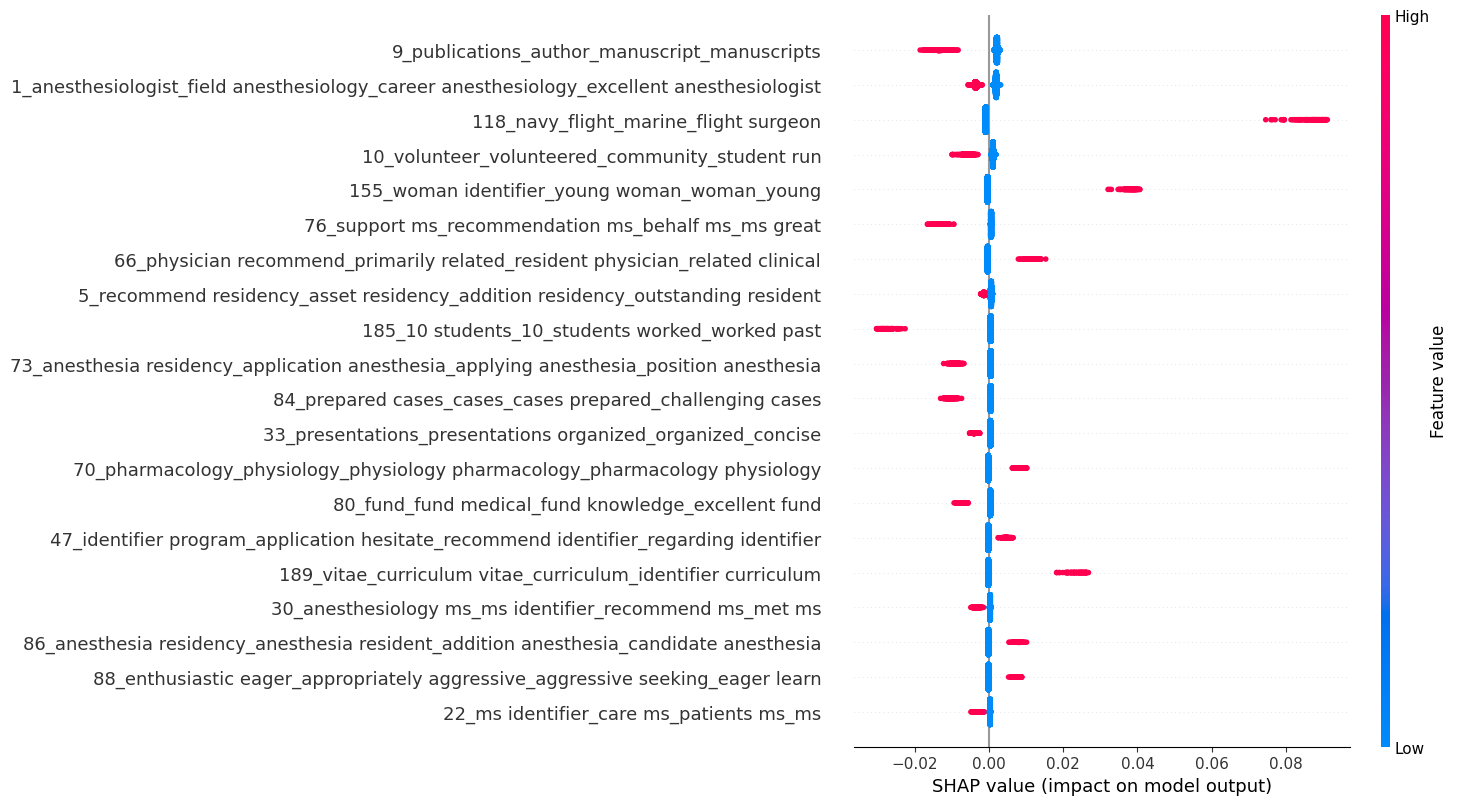}
  \caption{SHAP summary plot showing the most influential topics in predicting gender from recommendation letters using a Random Forest classifier trained on \textit{BERTopic} vectors. Each row represents a topic, labeled with its \textit{BERTopic} ID followed by the top five unigrams or bigrams (joined with underscores) extracted via TF-IDF. Each dot corresponds to a single letter; the horizontal position indicates the SHAP value, reflecting the topic's contribution to the prediction (positive values push toward "\textit{male}" negative toward "\textit{female}"). Red dots represent the presence of a topic in a letter, while blue dots represent its absence.}

  \label{fig:random_forest_classifier_shap_plot}
\end{figure}

To better understand which topics were most predictive of gender, SHAP was used to interpret the random forest model. Figure \ref{fig:random_forest_classifier_shap_plot} displays the top contributing topics and their relative impact on the model's predictions. Topic labels on the left were generated by aggregating all sentences assigned to each topic into a single document, then applying TF-IDF to extract the five most distinctive unigrams and bigrams. These are separated by underscores for readability. The number at the start of each label corresponds to the topic ID assigned by \textit{BERTopic}.

Each dot in the plot represents a single recommendation letter. The horizontal position of the dot reflects the SHAP value, which quantifies how much that topic influenced the model's prediction for that letter. Positive SHAP values indicate a push toward predicting "\textit{male}", while negative values indicate a push toward "\textit{female}". The dot color reflects the feature value: since our input features are binary topic vectors, red dots represent letters in which the topic was present, and their position indicates the strength and direction of its influence. Blue dots represent letters where the topic was absent.

\subsection{Topic and Text Embedding Infusion}

In addition to using topic vectors alone to predict the applicant's gender, the contextualized embedding representations produced by our fine-tuned DistilBERT model were also concatenated with them during training. This fusion allowed the classifier to leverage both the high-level thematic structure captured through topic modeling and the nuanced contextual signals encoded by the transformer. An 80:10:10 train, validation, and test split,  was used and the model was trained using the same hyperparameters outlined in Table~\ref{tab:best-distilbert-hparams}. Final evaluation results on the held-out test set are reported in Table~\ref{tab:topic-and-bert-performance}.

\section{Analysis of TF-IDF Tokens}
\label{sec:tf-idf_analysis}
For each gender-specific document, the top 10 tokens were extracted for each part of speech (adjectives, verbs, and nouns). The top 10 adjectives for male and female letters are shown in \autoref{tab:tfidf_male_adjectives} and \autoref{tab:tfidf_female_adjectives}, respectively. The top verbs are presented in \autoref{tab:tfidf_male_verbs} and \autoref{tab:tfidf_female_verbs}, and the top nouns in \autoref{tab:tfidf_male_nouns} and \autoref{tab:tfidf_female_nouns}.

\begin{table}[!htbp]
 \begin{minipage}{.5\linewidth}
\centering
\caption{Male Adjectives}
\small
\begin{tabular}{lccc}
\toprule
\textbf{Token} & \textbf{Female} & \textbf{Male} & \textbf{Diff} \\
\midrule
calm        & 0.021404 & 0.031890 & 0.010486 \\
young       & 0.030375 & 0.040406 & 0.010310 \\
medical     & 0.609714 & 0.619496 & 0.009782 \\
good        & 0.144768 & 0.154044 & 0.009275 \\
professional& 0.064924 & 0.074169 & 0.009246 \\
ethic       & 0.055420 & 0.064632 & 0.009211 \\
long        & 0.057019 & 0.065654 & 0.008635 \\
internal    & 0.056397 & 0.064164 & 0.007766 \\
appropriate & 0.033128 & 0.040831 & 0.007703 \\
respectful  & 0.016786 & 0.024482 & 0.007696 \\
great       & 0.198324 & 0.205860 & 0.007536 \\
able        & 0.143258 & 0.150765 & 0.007507 \\
critical    & 0.059950 & 0.067016 & 0.007066 \\
humble      & 0.014122 & 0.021033 & 0.006911 \\
interested  & 0.029487 & 0.035552 & 0.006065 \\
right       & 0.038190 & 0.044025 & 0.005834 \\
willing     & 0.018029 & 0.023801 & 0.005771 \\
inpatient   & 0.024513 & 0.030145 & 0.005632 \\
happy       & 0.030908 & 0.036446 & 0.005538 \\
hard        & 0.035348 & 0.040746 & 0.005398 \\
\bottomrule
\end{tabular}
\label{tab:tfidf_male_adjectives}
\end{minipage}
\begin{minipage}{.5\linewidth}
\centering
\caption{Female Adjectives}
\small
\begin{tabular}{lccc}
\toprule
\textbf{Token} & \textbf{Female} & \textbf{Male} & \textbf{Diff} \\
\midrule
identifi      & 0.404463 & 0.382980 & -0.021483 \\
outstanding   & 0.126650 & 0.111211 & -0.015439 \\
clinical      & 0.320622 & 0.308854 & -0.011768 \\
pediatric     & 0.043608 & 0.033721 & -0.009887 \\
public        & 0.013766 & 0.007025 & -0.006741 \\
global        & 0.011546 & 0.005663 & -0.005883 \\
identifier    & 0.053733 & 0.048580 & -0.005153 \\
numerous      & 0.028154 & 0.023204 & -0.004950 \\
new           & 0.057019 & 0.052242 & -0.004777 \\
social        & 0.017408 & 0.012731 & -0.004677 \\
future        & 0.051513 & 0.046962 & -0.004550 \\
compassionate & 0.036858 & 0.032529 & -0.004329 \\
academic      & 0.096187 & 0.092222 & -0.003965 \\
pre           & 0.018385 & 0.014434 & -0.003951 \\
bright        & 0.035260 & 0.031337 & -0.003923 \\
efficient     & 0.015187 & 0.011496 & -0.003692 \\
competitive   & 0.013056 & 0.009367 & -0.003689 \\
specific      & 0.015099 & 0.011751 & -0.003347 \\
warm          & 0.016075 & 0.012731 & -0.003345 \\
fantastic     & 0.013322 & 0.010048 & -0.003274 \\
\bottomrule
\end{tabular}
\label{tab:tfidf_female_adjectives}
\end{minipage}
\end{table}

\begin{table}[!htbp]
 \begin{minipage}{.5\linewidth}
\centering
\caption{Male Verbs}
\small
\begin{tabular}{lccc}
\toprule
\textbf{Token} & \textbf{Female} & \textbf{Male} & \textbf{Diff} \\
\midrule
show      & 0.105123 & 0.121443 & 0.016321 \\
like      & 0.046116 & 0.058660 & 0.012544 \\
feel      & 0.104389 & 0.114893 & 0.010504 \\
believe   & 0.092022 & 0.100338 & 0.008316 \\
learn     & 0.221565 & 0.228380 & 0.006815 \\
display   & 0.034482 & 0.041047 & 0.006565 \\
know      & 0.153859 & 0.160210 & 0.006351 \\
spend     & 0.071375 & 0.077291 & 0.005916 \\
supervise & 0.021800 & 0.026783 & 0.004982 \\
build     & 0.016140 & 0.020863 & 0.004723 \\
enjoy     & 0.046221 & 0.050800 & 0.004579 \\
ask       & 0.116652 & 0.120910 & 0.004258 \\
benefit   & 0.005764 & 0.009801 & 0.004036 \\
maintain  & 0.021171 & 0.024890 & 0.003719 \\
begin     & 0.022534 & 0.026152 & 0.003618 \\
attend    & 0.070117 & 0.073652 & 0.003535 \\
write     & 0.226701 & 0.230223 & 0.003523 \\
read      & 0.049574 & 0.052934 & 0.003360 \\
answer    & 0.017713 & 0.021009 & 0.003296 \\
require   & 0.049889 & 0.053128 & 0.003240 \\
\bottomrule
\end{tabular}
\label{tab:tfidf_male_verbs}
\end{minipage}
 \begin{minipage}{.5\linewidth}
\centering
\caption{Female Verbs}
\small
\begin{tabular}{lccc}
\toprule
\textbf{Token} & \textbf{Female} & \textbf{Male} & \textbf{Diff} \\
\midrule
take       & 0.133840 & 0.123918 & -0.009922 \\
complete   & 0.102293 & 0.093642 & -0.008651 \\
stand      & 0.038150 & 0.030276 & -0.007874 \\
support    & 0.052719 & 0.044977 & -0.007741 \\
organize   & 0.035530 & 0.027947 & -0.007583 \\
present    & 0.078292 & 0.071420 & -0.006872 \\
excel      & 0.065715 & 0.059096 & -0.006619 \\
match      & 0.034377 & 0.028432 & -0.005945 \\
care       & 0.054291 & 0.048568 & -0.005723 \\
recruit    & 0.029032 & 0.023483 & -0.005549 \\
include    & 0.115289 & 0.109896 & -0.005393 \\
hope       & 0.038779 & 0.033430 & -0.005349 \\
shadow     & 0.017608 & 0.012275 & -0.005332 \\
evaluate   & 0.028403 & 0.023144 & -0.005259 \\
try        & 0.021591 & 0.016448 & -0.005143 \\
run        & 0.020857 & 0.015963 & -0.004894 \\
waive      & 0.081960 & 0.077291 & -0.004669 \\
manage     & 0.036683 & 0.032217 & -0.004466 \\
encourage  & 0.014044 & 0.009655 & -0.004389 \\
reach      & 0.023896 & 0.019699 & -0.004198 \\
\bottomrule
\end{tabular}
\label{tab:tfidf_female_verbs}
\end{minipage}
\end{table}

\begin{table}[!htbp]
 \begin{minipage}{.5\linewidth}
\centering
\caption{Male Nouns}
\small
\begin{tabular}{lccc}
\toprule
\textbf{Token} & \textbf{Female} & \textbf{Male} & \textbf{Diff} \\
\midrule
medicine        & 0.151597 & 0.165628 & 0.014031 \\
staff           & 0.047931 & 0.059773 & 0.011843 \\
knowledge       & 0.114200 & 0.125307 & 0.011107 \\
physician       & 0.063732 & 0.072574 & 0.008842 \\
practice        & 0.024229 & 0.031521 & 0.007292 \\
time            & 0.136849 & 0.143022 & 0.006173 \\
anesthesia      & 0.105677 & 0.111843 & 0.006166 \\
year            & 0.201395 & 0.207183 & 0.005789 \\
number          & 0.028825 & 0.034538 & 0.005713 \\
rotation        & 0.166823 & 0.172302 & 0.005479 \\
demeanor        & 0.014796 & 0.020069 & 0.005273 \\
base            & 0.019392 & 0.024618 & 0.005225 \\
residency       & 0.225863 & 0.230636 & 0.004773 \\
training        & 0.079150 & 0.083820 & 0.004670 \\
week            & 0.046255 & 0.050836 & 0.004581 \\
discussion      & 0.018770 & 0.023269 & 0.004499 \\
question        & 0.089732 & 0.094151 & 0.004419 \\
resident        & 0.153272 & 0.157651 & 0.004378 \\
topic           & 0.017046 & 0.020983 & 0.003937 \\
anesthesiologist & 0.059422 & 0.063293 & 0.003871 \\
\bottomrule
\end{tabular}
\label{tab:tfidf_male_nouns}
\end{minipage}
 \begin{minipage}{.5\linewidth}
\centering
\caption{Female Nouns}
\small
\begin{tabular}{lccc}
\toprule
\textbf{Token} & \textbf{Female} & \textbf{Male} & \textbf{Diff} \\
\midrule
health        & 0.047691 & 0.032527 & -0.015164 \\
research      & 0.121909 & 0.108163 & -0.013746 \\
identifier    & 0.356199 & 0.342753 & -0.013446 \\
student       & 0.314541 & 0.305929 & -0.008612 \\
applicant     & 0.024851 & 0.019063 & -0.005788 \\
surgery       & 0.063780 & 0.058059 & -0.005721 \\
education     & 0.034571 & 0.029075 & -0.005496 \\
community     & 0.034380 & 0.029189 & -0.005190 \\
leadership    & 0.032991 & 0.028001 & -0.004990 \\
patient       & 0.342936 & 0.338182 & -0.004754 \\
child         & 0.014604 & 0.010012 & -0.004592 \\
clerkship     & 0.050947 & 0.046424 & -0.004523 \\
meeting       & 0.019105 & 0.014629 & -0.004476 \\
study         & 0.027293 & 0.022995 & -0.004298 \\
skill         & 0.141541 & 0.137261 & -0.004280 \\
passion       & 0.022792 & 0.018515 & -0.004277 \\
department    & 0.053820 & 0.049762 & -0.004059 \\
team          & 0.142882 & 0.138999 & -0.003883 \\
care          & 0.164429 & 0.160873 & -0.003556 \\
project       & 0.052240 & 0.048733 & -0.003507 \\
\bottomrule
\end{tabular}
\label{tab:tfidf_female_nouns}
\end{minipage}
\end{table}

\section{Computational Frameworks}

We used several standard software libraries in our experiments. All model loading and tokenization were performed using the HuggingFace Transformers library \cite{wolf-etal-2020-transformers}, and model fine-tuning was carried out using PyTorch \cite{paszke2019pytorch}. For evaluation, we used scikit-learn \cite{pedregosa2011scikit}, including its implementations of standard metrics such as accuracy and $F_1$ score. For model interpretability, we applied SHAP \cite{NIPS2017_8a20a862} to compute token-level attributions.

\section{Artifact Licenses}

Our experiments leverage the \textbf{Llama2-7B-chat} large language model \cite{touvron2023llama}. This model was accessed via the Hugging Face \texttt{transformers} library and is distributed under the \href{https://github.com/facebookresearch/llama/blob/main/LICENSE}{Llama 2 Community License}. We acknowledge and adhere to the terms of this license for our research. Further details on Llama2 are available at \href{https://ai.meta.com/llama/}{Meta AI's official Llama webpage}.

\end{document}